\documentclass[conference]{IEEEtran}
\IEEEoverridecommandlockouts
\usepackage{cite}
\usepackage{amsmath,amssymb,amsfonts}
\usepackage{algorithmic}
\usepackage{graphicx}
\usepackage{textcomp}
\usepackage{xcolor}
\usepackage{tabularx}
\usepackage{url}

\def\BibTeX{{\rm B\kern-.05em{\sc i\kern-.025em b}\kern-.08em
    T\kern-.1667em\lower.7ex\hbox{E}\kern-.125emX}}

\newcommand\system{MAA}

\newcommand\eg{{\it e.g.}}
\newcommand\fig{{Fig.~}}

\begin{document}

\title{Fine-Grained Scene Image Classification with Modality-Agnostic Adapter
}


\author{\IEEEauthorblockN{Yiqun Wang$^{1,2}$, Zhao Zhou$^{1,3}$, Xiangcheng Du$^{1,3}$, Xingjiao Wu$^{1}$, Yingbin Zheng$^{3}$, Cheng Jin$^{1,2*}$\thanks{$^{*}$Corresponding author.}}
    \IEEEauthorblockA{$^1$\textit{School of Computer Science, Fudan University, Shanghai, China}\\
    $^2$\textit{Innovation Center of Calligraphy and Painting Creation Technology, MCT, China}\\
    $^3$\textit{Videt Technology, Shanghai, China}\\
    \{yiqunwang23, zzhou21, xcdu22\}@m.fudan.edu.cn, \\\{xjwu\_cs, jc\}@fudan.edu.cn, zyb@videt.cn
    }    
}

\maketitle

\begin{abstract}
When dealing with the task of fine-grained scene image classification, most previous works lay much emphasis on global visual features when doing multi-modal feature fusion. In other words, models are deliberately designed based on prior intuitions about the importance of different modalities. In this paper, we present a new multi-modal feature fusion approach named \system~(Modality-Agnostic Adapter), trying to make the model learn the importance of different modalities in different cases adaptively, without giving a prior setting in the model architecture. More specifically, we eliminate the modal differences in distribution and then use a modality-agnostic Transformer encoder for a semantic-level feature fusion. Our experiments demonstrate that \system~achieves state-of-the-art results on benchmarks by applying the same modalities with previous methods. Besides, it is worth mentioning that new modalities can be easily added when using \system~and further boost the performance. Code is available at https://github.com/quniLcs/MAA.
\end{abstract}


\section{Introduction}
\label{sec:intro}

Different from base-class image classification, fine-grained image classification requires the model to distinguish a base class of images in a more subtle manner~\cite{welinder2010caltech,khosla2011novel,krause20133d}, and therefore requires a stronger feature embedding capability.
Among different base classes, fine-grained scene image classification~\cite{karaoglu2013text, wang2022knowledge} handles scene images that contain more kinds of information than single object images. 
Since many images from the same class are visually similar, different types of features, including the visual clues and the scene texts from the image, are employed to make use of all kinds of information for fine-grained scene image classification. Recent researches mainly focus on using better multi-modal feature embedding to improve the model performance \cite{bai2018integrating, mafla2020fine, wang2022knowledge}.

\begin{figure}[t]
	\centering
	\includegraphics[width=\linewidth]{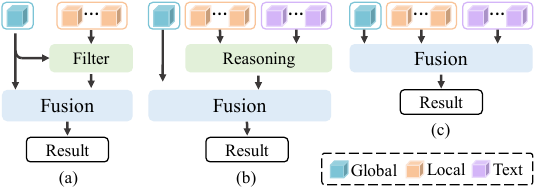}
	\caption{Different multi-modal feature fusion strategy for fine-grained scene image classification. Global (local) refers to global (local) visual embeddings, and text refers to text embeddings. While previous methods (a,b) lay much emphasis on global visual features, ours treats all modalities equally (c).}
	\label{fig:compare}
\end{figure}

\begin{figure}[t]
	\centering
	\includegraphics[width=\linewidth]{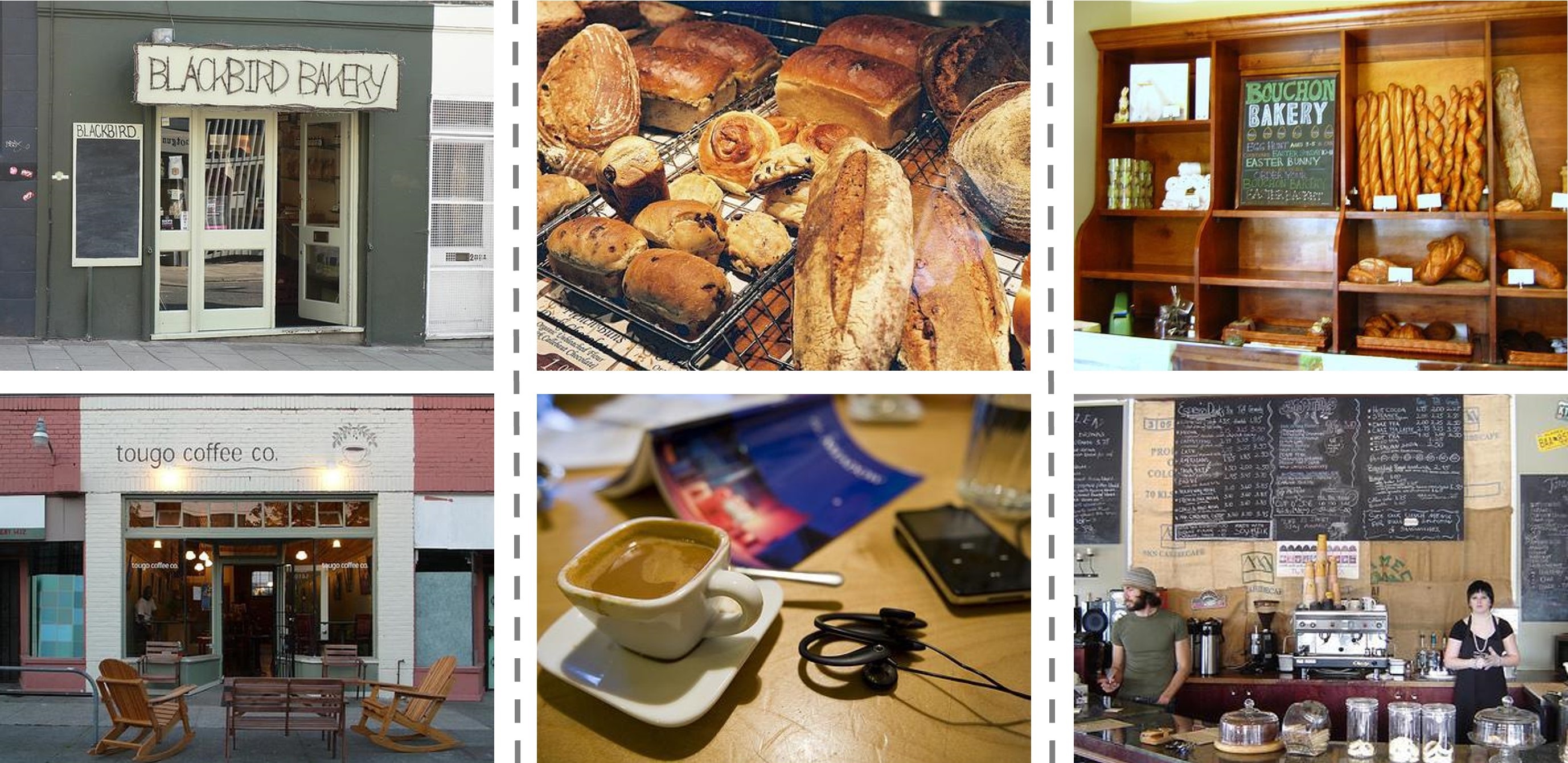}
	\caption{Example images from Con-Text dataset \cite{karaoglu2013text}. Images in the first and second rows are from class \textit{Bakery} and \textit{Cafe}, respectively. Images in the left column can be classified based on the text in the image, the middle column based on the object in the image, and the right column based on both clues.}
	\label{fig:intro}
\end{figure}

With these multi-modal embeddings, models continue with multi-modal feature fusion. As shown in \fig\ref{fig:compare}(a,b), most of them use the global visual feature vector as the query in the attention mechanism \cite{bai2018integrating, mafla2020fine, wang2022knowledge}, or concat the single global visual feature vector with the reasoning result of other modalities by methods like Graph Convolutional Network (GCN) \cite{mafla2021multi, li2022ga}. However, these operations consistently lay much emphasis on global visual features, based on a prior intuition that the global visual feature is always the most distinguishable and representative one among all modalities, which is actually not the case. As shown in \fig\ref{fig:intro}, texts and objects in the images can also be quite important, especially when the images have a similar global layout. For example, images in the first column both show a storefront and are hard to distinguish without using the names of store. According to images in different columns, we argue that the importance of different modalities differs in different cases, and there is no concrete modality preference in this task.

In this paper, we present \emph{Modality-Agnostic Adapter} (\system), a new multi-modal feature fusion approach for fine-grained scene image classification.
Our approach treats all modalities equally without a prior preference in the model architecture (\fig\ref{fig:compare}(c)).
Specifically, we first eliminate the modal differences in distribution and then use a modality-agnostic Transformer encoder for feature fusion, inspired by previous work~\cite{chen2020uniter, kim2021vilt, zhu2022uni, xu2021e2e}. In this way, the model can learn the importance of different modalities in different cases adaptively, instead of being given a prior setting within the model architecture. 
Our contributions are summarized as follows:
\begin{itemize}
	\item The proposed Modality-Agnostic Adapter treats all modalities equally for a better multi-modal feature fusion. Instead of designing a new fusion approach for each modality combination, it is easy to add new modalities and further improve the model performance. 
\item We achieve the state-of-the-art performance on fine-grained scene image classification benchmarks.
\end{itemize}


\section{Related Work}
\label{sec:related}


\noindent\textbf{Fine-grained scene image classification} requires the model to distinguish scene images, such as buildings \cite{karaoglu2013text} and crowd activities \cite{wang2022knowledge}, in a quite subtle manner. Since those images are visually similar, models always do multi-modal feature embedding first to make use of all kinds of information besides global visual features. For example, GoogLeNet \cite{szegedy2015going}, ResNet \cite{he2016deep} and ViT \cite{dosovitskiy2020image} models are used to get global visual embeddings; Faster R-CNN model pre-trained on the Visual Genome dataset \cite{ren2015faster, anderson2018bottom} is used to get salient region embeddings; E2E-MLT \cite{buvsta2019e2e} and Google OCR \cite{fujii2018optical} are used to spot texts in the images; 
Word2Vec \cite{mikolov2013efficient}, FastText \cite{bojanowski2017enriching}, and KnowBert \cite{peters2019knowledge} can be used to get word embeddings.

With these multi-modal feature embeddings, models continue with feature fusion. Generally speaking, most previous models are based on a prior intuition that the global visual feature is always the most distinguishable and representative one among all modalities \cite{bai2018integrating, mafla2020fine, mafla2021multi, wang2022knowledge, li2022ga}. For example, one previous work \cite{mafla2021multi} first uses a GCN to do semantic relation reasoning on the salient region embeddings and the word embeddings, and then concats the single global visual feature vector with the reasoning result; the state-of-the-art model \cite{wang2022knowledge} uses the global visual feature vector as the query in the attention mechanism and designs several residual operations differently for different modalities in the VKAC module. A more recent work \cite{zhang20232s} uses global and local visual embeddings as the query while text embeddings as the key and value in the attention mechanism, paying more attention to visual modality than to text modality. In this paper, we argue that the importance of different modalities differs in different cases, so we treat all modalities equally using a modality-agnostic Transformer encoder, which is the main contribution of this paper.


\vspace{0.1in}
\noindent\textbf{Multi-modal feature fusion}. 
Multi-modal representation learning is a more general problem, where multi-modal data not only refers to raw images, texts, audios and videos (called narrow multi-modal data), but also means multiple descriptors extracted from the raw data (called generalized multi-modal data) \cite{liang2021af}.
For narrow multi-modal data, quite a few vision-language models input images and texts to a single Transformer encoder to handle tasks like VQA (Visual Question Answering) and image-text retrieval \cite{chen2020uniter, kim2021vilt}, and this method can be extended to more modalities \cite{zhu2022uni} and Transformer decoder \cite{xu2021e2e}.  
For generalized multi-modal data, in the field of video retrieval, the method of extracting different modalities from raw videos has been explored in previous works like CE \cite{liu2019use} and MMT \cite{gabeur2020multi}.
However, the idea of treating all modalities equally is not common in the field of fine-grained image classification.




\begin{figure*}[t]
	\centering
	\includegraphics[width=\linewidth]{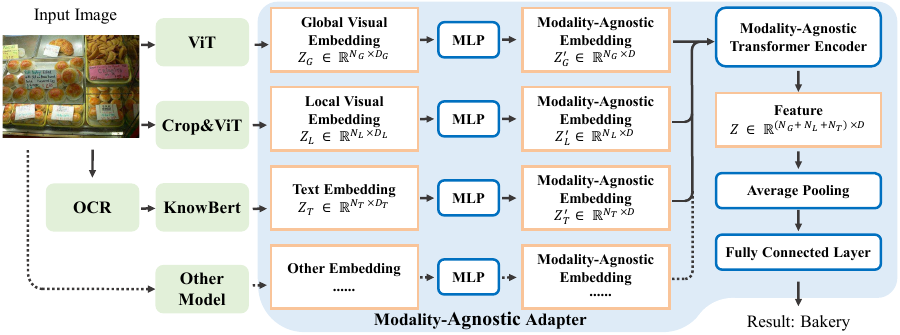}
	\caption{The architecture of \system. After getting the multi-modal feature embeddings (\eg, global image embedding, local region embedding and text embedding), several multi-layer perceptrons (MLPs) are used to eliminate the modal differences. Then the modality-agnostic embeddings are sent into the modality-agnostic Transformer encoder for a semantic-level fusion.}
	\vspace{0.10in}
	\label{fig:pipeline}
\end{figure*}


\section{Approach}
\label{sec:method}

The overall architecture of our proposed model is shown in Fig. \ref{fig:pipeline}, where multi-modal feature fusion is improved by using a modality-agnostic Transformer, and new modalities can further boost the model performance. 

\subsection{Overall Pipeline}
\label{subsec:overall}

Given an input image $I \in \mathbb{R}^{H \times W \times 3}$, different pre-trained models are first used to get multi-modal feature embeddings, denoted as $ Z_m \in \mathbb{R}^{N_m \times D_m} $, where $N_m$ stands for the sequence length of a single modality, $D_m$ stands for the dimension of feature, and $ m \in \{G, L, T\}$ stands for global visual feature, local visual feature and text feature respectively. 

Before going through the modality-agnostic Transformer, several multi-layer perceptrons (MLPs) are used to eliminate the modal differences in distribution, namely changing the dimensions of feature to the same, denoted as $D$. Then the obtained modality-agnostic embeddings $ Z_m' \in \mathbb{R}^{N_m \times D} $ can be concated and input into the modality-agnostic Transformer encoder for a semantic-level fusion. Finally, given the feature sequence $ Z \in \mathbb{R}^{(N_G+N_L+N_T) \times D} $, an average pooling operation is used to get a single feature vector $ z \in \mathbb{R}^D $ for each input image, while a fully connected layer is used to get the probability distribution of the classification result. Cross entropy loss is used during training:
\begin{equation}
\mathcal{L}_{Class} = -\log \dfrac{\exp(W_y^T z + b_y)}{\sum_{j=1}^C \exp(W_j^T z + b_j)}
\end{equation}
where $y$ is the index of the target category, $C$ is the number of classes in the dataset, $ W \in \mathbb{R}^{D \times C}$ and $b \in \mathbb{R}^C$ are the parameters of the fully connected layer.

\subsection{Multi-modal Feature Embeddings}

Inspired by previous work \cite{wang2022knowledge}, we use the output of ViT \cite{dosovitskiy2020image} as the global visual embedding and KnowBert \cite{peters2019knowledge} as the text embedding. Unlike previous work \cite{mafla2021multi}, we do not use Faster R-CNN pre-trained on Visual Genome \cite{anderson2018bottom} to detect salient objects; instead, we crop the upper-left, upper-right, bottom-left, bottom-right and center part of the original image and input the 5 cropped images to the ViT model to get local visual embeddings, inspired by previous work \cite{wei2015hcp, wang2016beyond}. In this way, the branch of local image embedding becomes quite simple, since it does not contain extra parameters or another complicated pre-trained model. 

\subsection{Modality-Agnostic Adapter}
\label{subsec:fusion}

After obtaining multi-modal feature embeddings from different pre-trained models, this paper divides multi-modal feature fusion into two steps: modal difference elimination and modality-agnostic feature fusion.

\vspace{0.1in}
\noindent\textbf{Modal difference elimination.}
On the one hand, since the current feature embeddings belong to different modalities, the distribution discrepancy among them will hinder the feature fusion process if not eliminated. In other words, the feature fusion module will pay more attention to the modal difference instead of the semantic relationship because of the distribution discrepancy, so it is unreasonable to directly perform feature fusion \cite{liang2021attention}. On the other hand, feature embeddings of different modalities may have different dimensions, so several MLPs are required to eliminate the dimensional difference. 
Based on the above two reasons, at this stage, the feature embeddings of different modalities will go through respective MLPs towards the modality-agnostic feature embeddings, which can be formulated as: 
\begin{equation} \label{equ:mlp}
Z_m'=\texttt{Act}(\texttt{FC}(\texttt{LN}(Z_m)))
\end{equation}
where \texttt{Act}, \texttt{FC} and \texttt{LN} stand for activation function, fully connected layer and layer normalization respectively. The dimensions and modalities of all feature embeddings are finally consistent to be $D$. Ablation studies on this module can be found in Section \ref{subsec:ablation}.

\vspace{0.1in}
\noindent\textbf{Modality-agnostic feature fusion.}
The obtained modality-agnostic feature embeddings then go through a modality-agnostic Transformer encoder for feature fusion, which can be formulated as: 
\begin{align}
Z_0=\texttt{Concat}(Z_G',Z_L',Z_T') \\
Z_{i+1}=\texttt{Transformer}_i(Z_i)
\end{align}
Inspired by previous work \cite{ye2023deepsolo, kim2021vilt}, we do not use position embedding in the Transformer encoder; instead, we use modality embedding to pass on the information of modality source. Notice that the order of the embeddings from different modalities in Equation \ref{equ:mlp} does not influence the final computational result after the average pooling operation, which ensures the prior equality of different modalities. Moreover, since the modal differences in distribution have been eliminated in the former step, the feature fusion operation is able to pay more attention to the semantic-level relationships. In this way, our model can learn the importance of different information in different cases adaptively.

Besides, it is worth mentioning that \system~allows easy addition of new modalities. If different modalities are treated differently according to a prior intuition, once a new modality is added, a new model or at least a branch must be designed from scratch, which is quite complicated. On the contrary, since \system~treats all modalities equally, it is easy to add new modalities and further improve the model performance, which is another benefit brought by this idea.

\section{Experiments}
\label{sec:exp}


\subsection{Setup}

\noindent\textbf{Datasets.} 
\emph{Con-Text} \cite{karaoglu2013text} is a subset of the ImageNet dataset, which further classifies the \emph{Building} and \emph{Places of Business} categories in the ImageNet dataset into 28 fine-grained categories, such as \emph{bakery}, \emph{cafe} and \emph{pizzeria}.
There are in total 24255 images in this dataset, including 16179 in the training set and 8076 in the test set.
Notably, not all images in this dataset contain texts, so it is necessary to use more information besides texts for the task.
\emph{Crowd Activity} \cite{wang2022knowledge} includes in total 8785 images from 21 categories. These categories range from people's daily lives, such as \emph{Celebrate Christmas}, \emph{Celebrity Speech} and \emph{Graduation Ceremony}, to demonstration scenes, such as \emph{Appeal for Peace}, \emph{Brexit} and \emph{Racial Equality}.
Unlike Con-Text, all images in this dataset contain texts.
However, the content of this dataset involves more background knowledge.

\vspace{0.1in}
\noindent\textbf{Implementation details.} 
Modality-agnostic Transformer in \system~consists of 2 Transformer encoder layers, where the dimension of the input and output feature is 768, the dimension of the intermediate feature is 2048 and the number of heads in the multi-head self-attention is 8.
The model is trained on each dataset for 50 epochs with a batch size of 8 with the AdamW \cite{ilya2019decoupled} optimizer. The basic learning rate is $3\times10^{-5}$ with a linear warm-up at the beginning and a cosine annealing warm restart schedule \cite{loshchilov2016sgdr} afterwards. 
Data augmentation include random crop, random horizontal flip, random color jittering, and random grayscale.  
The code implementation is based on PyTorch~\cite{paszke2019pytorch}, where the code of the ViT model is based on the timm package~\cite{rw2019timm}.
All the experiments are conducted on a single NVIDIA Titan Xp GPU.

\subsection{Comparison with State-of-the-arts}

Table \ref{table:experiment} reports the fine-grained scene image classification results on the Con-Text and Crowd Activity datasets. Following previous methods, we use mAP as the metric.
For better comparison with previous models, we first use the output of ViT \cite{dosovitskiy2020image} as the global visual embedding and KnowBert \cite{peters2019knowledge} as the text embedding, without using local visual feature. In this way, both our model and the state-of-the-art \cite{wang2022knowledge} use the same pre-trained feature extractors, which ensures comparison fairness. From the results, our method \system~outperforms the state-of-the-art model on both datasets without using additional information, which demonstrates the effectiveness of our fusion approach. In other words, there is no need to deliberately design a model based on prior intuitions; instead, a modality-agnostic Transformer itself can learn the importance of different modalities adaptively. 

Besides global visual embedding and text embedding, other embeddings like local visual embedding can also be added to our model. As listed in Table \ref{table:experiment}, using a simple local visual embedding further boosts the performance. For example, on the Con-Text dataset, the mAP of our model reaches 90.13\% without using additional modalities, surpassing the state-of-the-art model by 0.60\%; adding local visual embedding further improves the mAP by 0.74\% to 90.87\%.

\begin{table}[t]
		\centering
		\caption{Comparison with state-of-the-art methods. } 
		\label{table:experiment}
		\begin{tabular}{l|ccc|cc}
			\hline
			Method & Global & Local & Text & Con-Text & Activity \\
			\hline
			Karaoglu \textit{et al.} \cite{karaoglu2013text} & \checkmark & & \checkmark & 39.00 & -\\
			Karaoglu \textit{et al.} \cite{karaoglu2016words} & \checkmark & & \checkmark & 77.30 & -\\
			Bai \textit{et al.} \cite{bai2018integrating} & \checkmark & & \checkmark & 79.60 & -\\
			Mafla \textit{et al.} \cite{mafla2020fine} & \checkmark &  & \checkmark & 80.21 & 77.57 \\
			Mafla \textit{et al.} \cite{mafla2021multi} & \checkmark & \checkmark & \checkmark & 85.81 & 79.25 \\
			GA-SRN \cite{li2022ga} & \checkmark & \checkmark & \checkmark & 86.57 & - \\
			2S-DFN \cite{zhang20232s} & \checkmark & \checkmark & \checkmark & 87.83 & - \\
			Wang \textit{et al.} \cite{wang2022knowledge} & \checkmark & & \checkmark & 89.53 & 87.45 \\
			\hline
			\system* & \checkmark & & \checkmark & \underline{90.13} & \underline{88.07} \\
			\system & \checkmark & \checkmark & \checkmark & \textbf{90.87} & \textbf{88.29} \\
			\hline
		\end{tabular}
		\\
		\vspace{0.06in}
		\textbf{Bold} and \underline{underlined} indicate the top and runner-up results, respectively. \\
		* means local embeddings are not used.
		
\end{table}

\begin{figure*}[t]
	\centering
	\includegraphics[width=\linewidth]{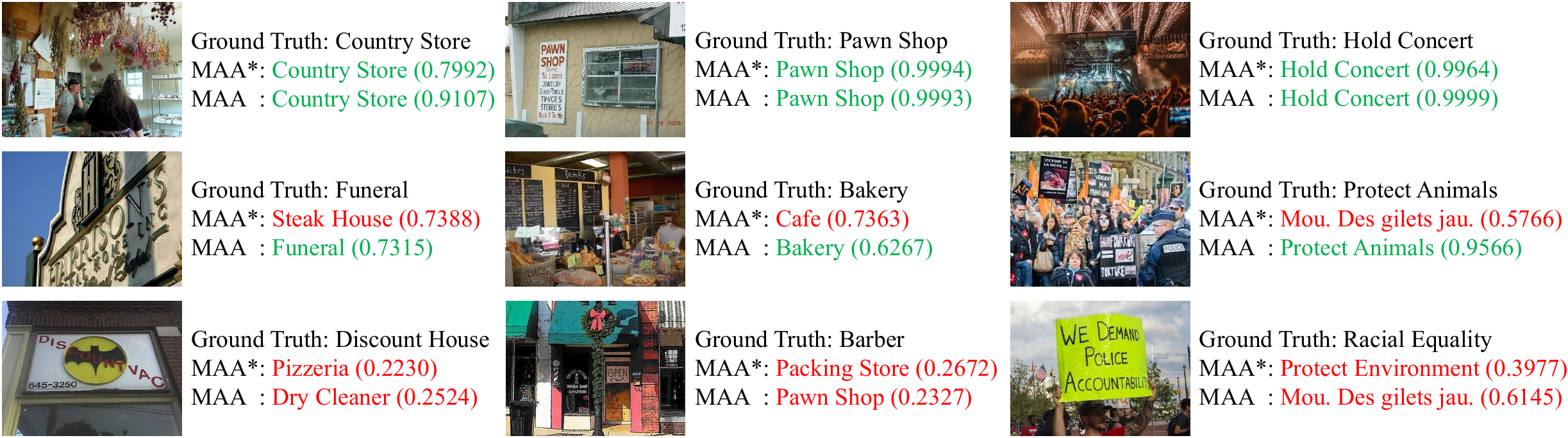}
	\caption{Some examples of model prediction. Images in the first and second columns are from Con-Text \cite{karaoglu2013text}, while the last column are from Crowd Activity \cite{wang2022knowledge}. Texts in green and red refer to correct and incorrect predictions respectively.}
	\label{fig:examples}
\end{figure*}

\subsection{Ablation Study}
\label{subsec:ablation}

\begin{table}[t]
	\centering
	\caption{Effects of different modalities.}
	\label{table:ablation}
	\begin{tabular}{p{30px}<{\centering} p{30px}<{\centering} p{30px}<{\centering}|p{48px}<{\centering} p{48px}<{\centering}}
		\hline
		Global & Local & Text & Con-Text & Activity \\ 
		\hline
		\checkmark & & & 82.17 & 79.65\\
		\checkmark & \checkmark & & 82.90 & 81.69 \\
		& \checkmark & \checkmark & 89.98 & 88.18\\
		\checkmark & & \checkmark & 90.13 & 88.07 \\
		\checkmark & \checkmark & \checkmark & \textbf{90.87} & \textbf{88.29} \\
		\hline
	\end{tabular}
\end{table}

\noindent\textbf{Different modalities.}
As shown in Table \ref{table:ablation}, adding a new modality can always contribute to the performance of our model. 
However, compared with global and local visual embeddings, adding text embedding leads to a more remarkable improvement, which does not accord with the prior intuition of previous works.
Besides, the relative importance of different modalities is not consistent between 2 datasets. For example, global visual embedding is slightly more important than local visual embedding for the Con-Text dataset \cite{karaoglu2013text}, which is not the case for the Crowd Activity dataset \cite{wang2022knowledge}, probably because the latter includes more local details.
Moreover, the interactions between different modalities are also quite complicated. For example, on the Con-Text dataset \cite{karaoglu2013text}, using local visual embedding brings an improvement on mAP of about 0.74\% whether there are text embeddings or not, while on the Crowd Activity dataset \cite{wang2022knowledge}, using local visual embedding only brings a minor improvement when text embedding already provides adequate information to the model.
These observations support our claim that the importance of different modalities differs in different cases, and that the model should not have a specific modality preference.  That is the reason why we let the model itself learn the importance of different modalities in different cases adaptively, without giving a prior setting in the model architecture.

\begin{table}[t]
	\centering
	\caption{Effects of modal difference elimination module.}
	\label{table:elimination}
	\begin{tabular}{p{110px}|p{45px}<{\centering} p{45px}<{\centering}}
		\hline
		Mode & Con-Text & Activity \\ 
		\hline
		None & 89.68 & 88.27 \\
		Shared MLPs & 90.16 & 88.22 \\
		Independent MLPs (\system) & \textbf{90.87} & \textbf{88.29} \\
		\hline
	\end{tabular}
\end{table}

\begin{table}[t]
	\centering
	\caption{Effects of Transformer layer numbers.}
	\label{table:layer}
	\begin{tabular}{p{56px}|p{30px}<{\centering} p{30px}<{\centering} p{30px}<{\centering} p{30px}<{\centering}}
		\hline
		\# Layers & 0 & 1 & 2 & 4 \\
		\hline
		Con-Text & 84.64 & 90.58 & \textbf{90.87} & 89.82\\
		Activity & 81.99 & 88.28 & \textbf{88.29} & 88.15\\
		\hline
	\end{tabular}
\end{table}

\vspace{0.1in}
\noindent\textbf{Modal difference elimination.}
Though modal difference elimination is necessary since feature embeddings of different modalities may have different dimensions, the output dimensions of ViT \cite{dosovitskiy2020image} and KnowBert \cite{peters2019knowledge} are coincidentally the same, which makes the ablation study feasible.
To ensure a fair comparison with the independent MLPs in \system, we also do experiments on the setting of shared MLPs. When using independent MLPs, the feature embeddings of different modalities will go through a respective single layer of MLP, whose parameters are not shared between modalities. In contrast, when using shared MLPs, the feature embeddings of different modalities will all go through the same multiple layers of MLPs, whose parameters are shared between modalities. In this way, both the mode of shared MLPs and independent MLPs have the same number of trainable parameters, and the mode of shared MLPs leads to a larger amount of computation.
As listed in Table \ref{table:elimination}, using independent MLPs to eliminate the modal differences can help the model to reach the best result, which demonstrates the necessity and effectiveness of this module.

\vspace{0.1in}
\noindent\textbf{Modality-agnostic Transformer encoder.} 
We stack 2 Transformer layers as the modality-agnostic Transformer encoder in \system. As shown in Table \ref{table:layer}, without the Transformer encoder, the mAP on both datasets drops about 7\%, demonstrating the importance of effective feature fusion. As the number of Transformer layers increases, the mAP also increases towards an upper bound, where the model may start to over-fit the training set. Therefore, we choose 2 as a suitable number of Transformer layers.





\subsection{Qualitative Analysis}

Fig. \ref{fig:examples} shows some examples of model prediction. 
As shown in the first row, adding local embedding helps raise the confidence score of the model prediction, especially when the local details in the image are relatively important.
In some cases, the local details are even decisive for a correct prediction. For example, when handling the image of \emph{Bakery} from Con-Text \cite{karaoglu2013text}, model with local embedding gives a correct prediction with the help of the baskets of bread at the bottom of the image, while model without local embedding gives a false prediction of \emph{Cafe} because of the drink menu; when handling the image of \emph{Protect Animals} from Crowd Activity \cite{wang2022knowledge}, model with local embedding gives a correct prediction with the help of the image of tortured animals on the posters, while model without local embedding gives a false prediction of \emph{Mou. Des gilets jau.} since there are French words on the posters.

As for the third row, it shows that one way to further improve the model performance is to use better embeddings. For example, Google OCR \cite{fujii2018optical} fails to spot the word `discount' in the image of \emph{Discount House} because of the complicated background; the barber's pole in the image of \emph{Barber} is very hard to detect even for the Faster R-CNN model pre-trained on Visual Genome \cite{anderson2018bottom}. Besides, the model actually over-fits the training set to some extent, since the model simply connects the yellow color with the yellow vests movement. Using better pre-trained models and embeddings in the future can also mitigate this problem.

\section{Conclusions}
\label{sec:conclusion}

In this paper, we presented a new multi-modal feature fusion approach \system~for fine-grained scene image classification, which can learn the importance of different modalities in different cases adaptively.
More specifically, we first eliminated the modal differences in distribution and then used a modality-agnostic Transformer encoder for a semantic-level feature fusion.
Since all modalities were treated equally in \system, new modalities can be easily added and further boost the model performance.
Experimental results on the Con-Text and Crowd Activity datasets demonstrated that both the proposed method and the added modalities contributed to the model performance.


\section*{Acknowledgment}

This work is supported by the National Natural Science Foundation of China (Grant No. 62207013) and the Shanghai Archives Research Program (Grant No. 2425 and 2108).

\small{
	\bibliographystyle{IEEEbib}
	\bibliography{total}
}

\end{document}